\documentclass[singleblind]{ceurart}

\usepackage{latexsym}
\usepackage{amssymb}
\usepackage{amsmath}
\usepackage{amsthm}
\usepackage{booktabs}
\usepackage{enumitem}
\usepackage{graphicx}
\usepackage{hyperref}
\usepackage{wrapfig}
\usepackage{float}

\usepackage{multirow}

\usepackage{blindtext}

\usepackage[ruled, lined, linesnumbered, commentsnumbered]{algorithm2e}
\usepackage{algpseudocode}

\usepackage{caption}
\usepackage{subcaption}

\usepackage{csquotes}

\usepackage{appendix}

\usepackage{cases}

\newcommand{\METR}{\textbf{METR}\xspace}
\newcommand{\METRpp}{\textbf{METR++}\xspace}
\newcommand{\TR}{\emph{Tree-Ring}\xspace}
\newcommand{\Stablesig}{\emph{Stable Signature}\xspace}
\newcommand{\nrm}{$\mathcal{N}$}
\newcommand{\bx}{\mathbf{x}}

\newcommand{\up}{$\mathbf{\uparrow}$}
\newcommand{\down}{$\mathbf{\downarrow}$}

\newcommand{\ie}{\emph{i.e.,}\xspace}
\newcommand{\eg}{\emph{e.g.,}\xspace}

\begin{document}


\conference{}

\title{METR: Image Watermarking with Large Number of Unique Messages}


\author[1,2]{Alexander Varlamov}
\address[1]{MIPT}
\address[2]{VK Lab}

\author[3]{Daria Diatlova}
\address[3]{deepvk, VK}

\author[2]{Egor Spirin}

\begin{abstract}
Improvements in diffusion models have boosted the quality of image generation, which has led researchers, companies, and creators to focus on improving watermarking algorithms. This provision would make it possible to clearly identify the creators of generative art. The main challenges that modern watermarking algorithms face have to do with their ability to withstand attacks and encrypt many unique messages, such as user IDs. In this paper, we present METR: Message Enhanced Tree-Ring, which is an approach that aims to address these challenges. METR is built on the Tree-Ring watermarking algorithm, a technique that makes it possible to encode multiple distinct messages without compromising attack resilience or image quality. This ensures the suitability of this watermarking algorithm for any Diffusion Model. In order to surpass the limitations on the quantity of encoded messages, we propose METR++, an enhanced version of METR. This approach, while limited to the Latent Diffusion Model architecture, is designed to inject a virtually unlimited number of unique messages. We demonstrate its robustness to attacks and ability to encrypt many unique messages while preserving image quality, which makes METR and METR++ hold great potential for practical applications in real-world settings. Our code is available at \url{https://github.com/deepvk/metr}.
\end{abstract}


\maketitle

\vspace{-0.5cm}
\section{Introduction}

Nowadays, image generation is one of the major applications of computer vision technology. It is used in various spheres, including entertainment~\cite{civitai, dalle}, medicine~\cite{medical-img-gen}, security~\cite{art-or-forgery}, and retail~\cite{virtual-try-on}. Recent advances in deep learning~\cite{lecun-deep-learn}, such as Variational Autoencoders (VAE)~\cite{vae}, Generative Adversarial Networks (GAN)~\cite{GAN}, and Diffusion models~\cite{diffusion_ddpm}, have allowed it to become a rapidly growing area of research. The latest achievements in text-to-image models, namely DALL-E 2~\cite{dalle}, Kandinsky~\cite{kandinsky}, and Stable Diffusion~\cite{LDM}, facilitate the generation of highly realistic images based on specific prompts.

With the growing popularity of image generation, the risks of it being used inappropriately or maliciously are also increasing. 
These risks include policy~\cite{civitai, pinterest} and privacy violations~\cite{law_gen_ai}, the generation of fake news~\cite{ai_gen_news}, document fraud~\cite{id_card_generation, document_generation}, and the creation of harmful content~\cite{unsafe_generation}.
To help alleviate these risks, it is necessary to develop a mechanism to detect whether an image is generated or not.
One possible solution is to label generated images with special messages, watermarks~\cite{fourier_wm, svd_wm, wen2023treering}.
To ensure the suitability of this solution for practical applications, it is essential that these watermarks remain invisible~\cite{zhao2023invisible, wen2023treering}. Additionally, the watermarks should be robust to a range of attacks~\cite{zhao2023invisible, wm_survey}, ensuring that perturbations to the image cannot remove an encoded message.

\begin{figure}[ht]
\centering
    \begin{subfigure}[htp]{0.5\textwidth}
        \centering
        \includegraphics[width=0.99\textwidth]{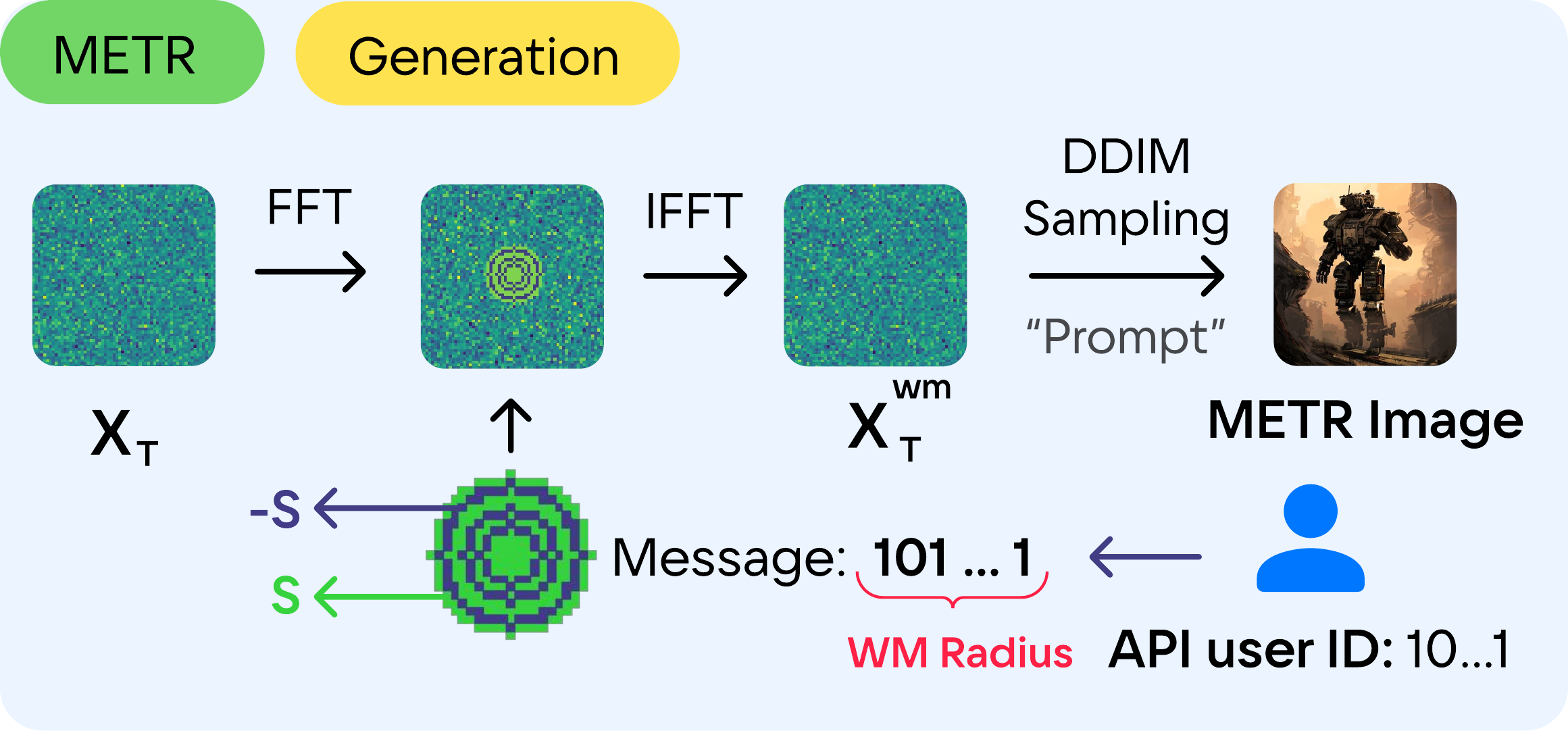}
        \subcaption{METR Generation process}
        \label{fig:metr-gen}
    \end{subfigure}%
    \begin{subfigure}[htp]{0.5\textwidth}
        \centering
        \includegraphics[width=0.99\textwidth]{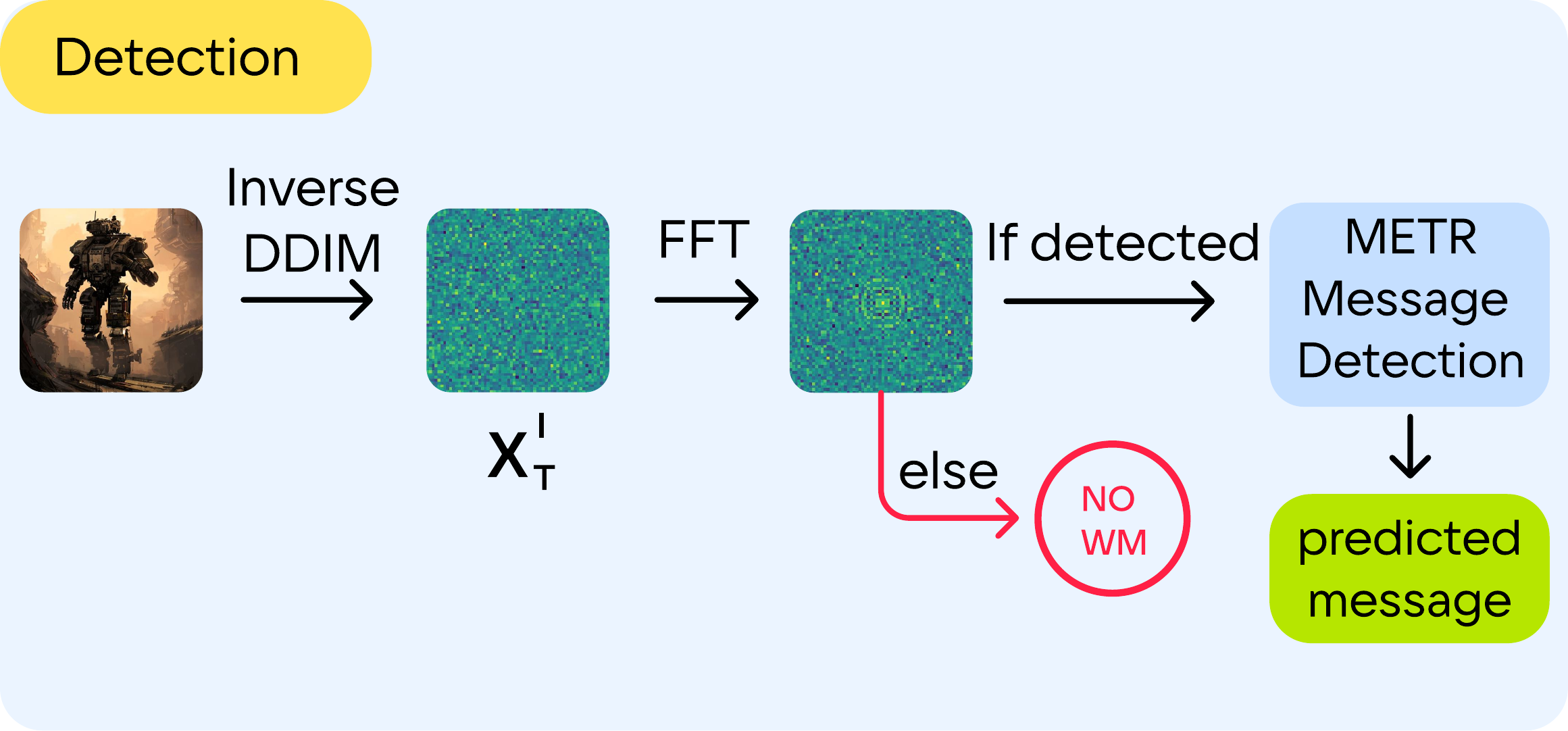}
        \subcaption{METR Detection process}
        \label{fig:metr-det}
    \end{subfigure}
\caption{\METR watermarking pipeline. Figure (a) outlines the steps to encrypt a binary message into an image via corresponding latent noise. Figure (b) details the process of detecting whether an image contains a watermark and deciphering the encrypted message.}
\label{fig:METR}
\vspace*{-5mm}
\end{figure}


Watermarking has been a widely used technique for image content protection long before the emergence of generative models.
The first works on the subject introduced algorithms that make it possible to add watermarks to existing images~\cite{fourier_wm, svd_wm, early_wm}. These methods can be used to apply watermarks to generated images as well.
However, this approach has several drawbacks. For example, the watermarks are not necessarily completely invisible to a human's eye~\cite{wen2023treering, fernandez2023stable}, and the latest algorithms~\cite{zhu2018hidden, rivagan, stegastamp} require training of additional model. Moreover, if the watermarking step is not built into the image generation process, then those with access to the generative model could generate images without watermarks.

A different group of algorithms was proposed, which add watermarks to images during the generation process~\cite{gan_wm, fernandez2023stable, liu2023watermarking}.
These approaches successfully address the challenges of previous algorithms. Newer watermarking methods require either tuning of the model's weights~\cite{gan_wm, fernandez2023stable, liu2023watermarking}, or adjustments to some latent image representation (\eg initial noise), that is used in the generation process~\cite{wen2023treering}. 
Both approaches make it possible to create \enquote{truly invisible} watermarks~\cite{wen2023treering}. Moreover, the second one is applicable to any diffusion architecture with a limitation of only sampling strategy (\eg DDIM~\cite{ddim}), and it also does not require additional model training. 
The only disadvantage of the second approach for watermarking images in a generation process can occur when someone gains access to the model's weights. In this scenario, the algorithm responsible for watermark embedding could be removed from the inference pipeline. But this drawback is fixed for the first approach, since the ability to watermark images is contained within the tuned weights of a model.
In a recent work by Wen~et.~al.~\cite{wen2023treering}, the authors present \TR, a watermarking algorithm for diffusion models that utilizes initial noise as a latent representation of the image. A watermark is injected as a subtle modification of the initial noise used during diffusion model sampling. 
This approach shows high robustness to any white-box attacks~\cite{zhao2023invisible, wm_survey}, \ie attacks only on the output of the model, as the model's weights stay inaccessible.


In addition to being able to determine whether an image is generated or not, it is also important to find out the author of the generated image.
This can be done by incorporating watermarks that contain specific information, such as user IDs.
Despite the high reliability and imperceptibility of \TR watermarks, the algorithm cannot encrypt messages within the watermark, limiting its practical use for certain real-world scenarios.
One of the existing algorithms capable of encrypting messages, \Stablesig~\cite{fernandez2023stable}, requires training a separate model for each user, as it can only manage one unique message per model. This is also not ideal for practical use in real-world applications.

In this paper, we propose \METR, a watermarking algorithm based on \TR watermarking~\cite{wen2023treering} of diffusion models~\cite{diffusion_ddpm, ddim}. This approach is able to handle many unique messages and is robust to white-box attacks~\cite{zhao2023invisible, wm_survey} without a loss in image quality.
Similar to \TR, \METR utilizes a modification of initial noise distribution without changing model architecture. Therefore, it does not require any additional training and can be easily transferred to another model.
In addition, we suggest a simple modification of \METR, \METRpp. It combines \METR with \Stablesig~\cite{fernandez2023stable} and extends the amount of unique messages encrypted to \emph{as many as one might need}. Our contributions can be summarized as follows:

\begin{itemize}
    \item \METR (Message Enhanced \TR) – a new watermarking algorithm that is capable of encoding a large number of unique messages into a \TR watermark without noticeable image quality degradation or a decrease in watermark robustness to attacks.
    \item We introduce an algorithm to select an optimal value of the hyperparameter $S$ for the \METR watermark based on the \enquote{detection resolution} metric, which captures the difference in detection distances between an image with and without a watermark.
    \item \METRpp – an extended version of the \METR watermarking algorithm, combines \METR with the \Stablesig algorithm and allows the encryption of an even larger number of unique messages compared to \METR.
\end{itemize} 




\vspace{-0.4cm}
\section{Related Works}
\label{sec:related-works}
\subsection{Diffusion Process}

Diffusion models~\cite{diffusion_ddpm,ddim} are generative models employed to approximate a data distribution $q(\bx_0)$ using a parametrized form, $p_\theta(\bx_0)$, which is expressed via latent variables $\bx_1,...,\bx_T$. The parameters $\theta$ are optimized to maximize the evidence lower bound (ELBO).

The forward diffusion process involves the gradual addition of noise to the initial data point $\bx_0$: $\bx_t=\sqrt{\bar{\alpha}_t} \bx_0+\sqrt{1-\bar{\alpha}_t} \epsilon$, where $\epsilon\sim \mathcal{N}(0, \text{I})$, $\bar{\alpha}_t$ are values that parametrize variance at step $t$ for distribution $q(\bx_t | \bx_{t-1})$.

A point from the initial distribution can be approximated by employing a denoising process. While reverse diffusion was characterized as probabilistic in~\cite{diffusion_ddpm}, a deterministic sampling method with an equivalent evidence lower bound (ELBO) was introduced in~\cite{ddim}, known as DDIM sampling: $\bx_0'^{(t)} := \mathbf{D}_\theta(\bx_t)= \cfrac{\bx_t-\sqrt{1-\bar{\alpha}_t}\cdot\epsilon_{\theta}(\bx_t, t)}{\sqrt{\bar{\alpha}_t}}$,
where $\epsilon_{\theta}$ is a trained model that predicts the noise added to $\bx_0$ to obtain $\bx_t$. Meanwhile, $\bx_0'^{(t)}$ represents an estimate of $\bx_0$ derived from the denoising of $\bx_t$.

The determinism of DDIM sampling~\cite{ddim} can be leveraged to trace back the initial noise from which the image was generated. This process is called \enquote{inverse diffusion}~\cite{wen2023treering}. We can obtain an estimate of the initial noise, $\bx_T'$, based on the assumption that: $\bx_{t+1}-\bx_{t} \approx \bx_{t}-\bx_{t-1}$. Each step of DDIM inversion mirrors a step of the forward process, but utilizes \enquote{trained noise}: $\bx_{t+1}=\sqrt{\bar{\alpha}_{t+1}} \bx_0'^{(t)}+\sqrt{1-\bar{\alpha}_{t+1}} \epsilon_\theta(\bx_t, t)$. $T$ steps of DDIM Inversion estimate the initial noise of an image: $\mathbf{D}^{inv}_\theta(\bx_0)=\bx_T'\approx\bx_T$.

\vspace{-0.3cm}
\subsection{Tree-Ring Watermark}
\label{sec:related-tree-ring}

The \TR watermarking method, proposed in~\cite{wen2023treering}, employs the DDIM inversion technique~\cite{ddim}. In this method, the watermark is encoded as concentric circles or squares within the Fourier space of the initial noise: $\mathcal{F}(\bx_T)$. This encoding results in a modified version of the initial noise, denoted as $\bx_T^\text{wm}$. Subsequently, DDIM sampling is applied to produce a watermarked image from the noise that carries the embedded fingerprint: $\bx_0^\text{wm}=\textbf{D}_\theta(\bx_T^\text{wm})$. 
For watermark detection, the DDIM Inversion process is used to estimate the initial noise and identify the watermark within the Fourier representation of the approximated initial noise of an image: $\text{WM}'=\mathcal{F}(\mathbf{D}^{inv}_\theta(\bx_0^\text{wm}))$.

The \TR watermark, as presented in~\cite{wen2023treering} transforms the distribution of $\bx_T$ into a non-Gaussian form.
Since $\mathcal{F}[e^{-ax^2}](q)\sim e^{-\pi^2q^2/a}$, we can determine whether a watermark is present or absent on the image by assessing if the distribution of $y=\mathcal{F}(\bx_T')$, where $\bx_T'$ is predicted initial noise in the inverse DDIM process, deviates from normality.
Non-normality can be assessed by performing a test on the null hypothesis $\mathcal{H}_0$: $y=\mathcal{F}(\bx_T')\sim \mathcal{N}(0, \sigma^2 I)$.  $\sigma$ can be estimated for every input: $\sigma\approx \cfrac{1}{M}\sum_{i=1}^{|M|} |y_i|^2$, where $M$ is a watermarked area of an image. To calculate p-value for $\mathcal{H}_0$, we can define $z(y)=\cfrac{1}{\sigma^2}\sum_{i=1}^{|M|}|\text{WM}_i-y_i|^2$, where WM denotes encrypted watermark values on the area $M$. We then apply the following equation:
\begin{equation}
\label{eq:p-value}
    p=\mathbb{P}(\chi^2_{|M|, \lambda}<z|\mathcal{H}_0)=F_{\chi^2}(z),
\end{equation}
where $\chi^2_{|M|, \lambda}$ denotes a non-central chi-squared random variable~\cite{noncentral-chi} with $\lambda=\cfrac{1}{\sigma^2}\sum_{i=1}^{|M|} |\text{WM}_i|^2$, and $F_{\chi^2}$ representing its cumulative distribution function~\cite{cdf-chi}. Large p-values indicate the presence of a watermark in the image, while small p-values indicate its absence.

\vspace{-0.3cm}
\subsection{Latent Diffusion and Stable Signature}


The Latent Diffusion Model concept, proposed in~\cite{LDM}, involves implementing the diffusion process within a latent space, which substantially enhances the quality of image generation.
A key architectural update is the incorporation of a Variational Autoencoder (VAE)~\cite{vae} model, which converts the original image into a compact, low-dimensional representation for use in the subsequent diffusion process. During inference, noise is sampled and then transformed into a latent representation using a trained diffusion model. This representation is finally decoded back into the end image through the decoder component of the VAE.

\begin{figure}[h]
\centering
    \includegraphics[width=0.95\linewidth]{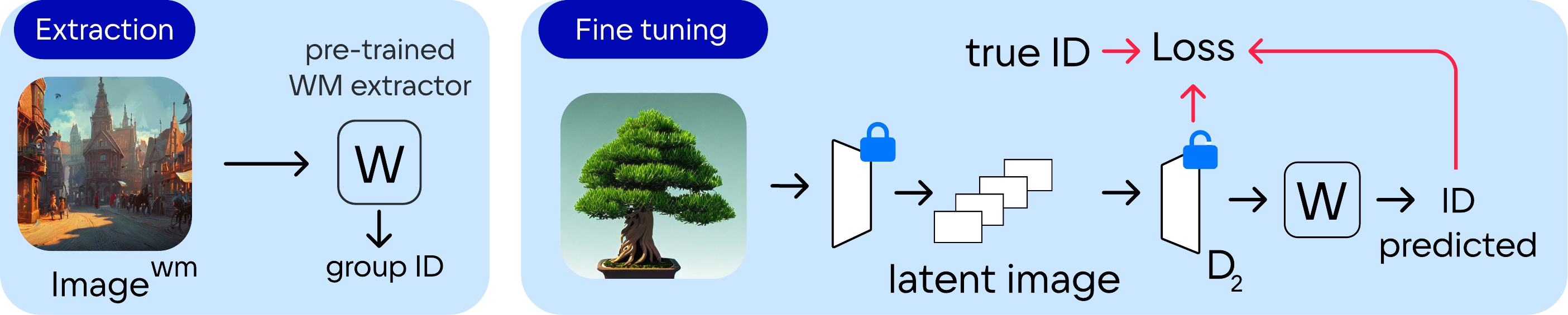}
    \caption{Stable-Signature~\cite{fernandez2023stable} scheme. Algorithms for watermark extraction and VAE decoder fine-tuning.}
\label{fig:stable-sig-scheme}
\vspace*{-5mm}
\end{figure}

Drawing on the Latent Diffusion Model concept that leverages VAE, the \Stablesig watermarking method was proposed in~\cite{fernandez2023stable}. \Stablesig enables the encryption of binary messages into images during their generation. To embed a \Stablesig key into an image, the decoder weights of the Variational Autoencoder (VAE) in the latent diffusion model are fine-tuned. This fine-tuning incorporates a loss function that includes an extra term for message detection: $L=L_\text{message}+\lambda L_\text{image}$. Message detection is carried out by the pre-trained network $W$, as detailed in~\cite{zhu2018hidden}. During the inference process, \Stablesig relies solely on the $W$ network to detect and retrieve the watermark. For further information on the Fine-Tuning and Extraction procedures, refer to Figure~\ref{fig:stable-sig-scheme}.

The inference process for \Stablesig is simple, yet the watermarking algorithm requires a unique Variational Autoencoder (VAE) decoder to be trained for each specific message. In real-world scenarios, especially for services with millions of users where messages are often user IDs, the implementation of \Stablesig is not feasible.



\vspace{-0.3cm}
\subsection{Watermark Attacks}
\label{sec:related-attacks}


\begin{wrapfigure}[11]{r}{0.45\textwidth} 
    \centering
    \begin{subfigure}[b]{0.21\textwidth}
        \centering
        \includegraphics[width=0.8\textwidth]{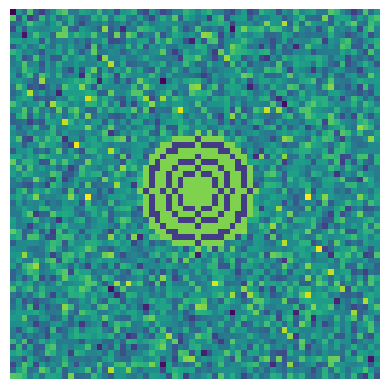}
        \caption{encrypted message}
    \end{subfigure}
    \hspace{0.01\textwidth}
    \begin{subfigure}[b]{0.21\textwidth}
        \centering
        \includegraphics[width=0.8\textwidth]{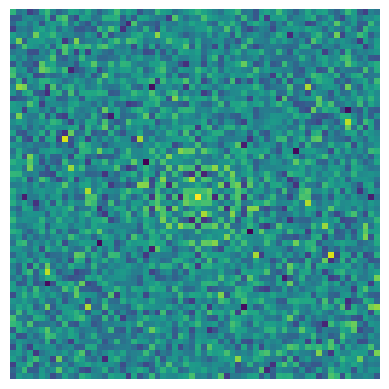}
        \caption{decrypted message}
    \end{subfigure}
    \caption{Message becomes circles in Fourier space $\mathcal{F}$ of the latent noise $\bx_T$}
    \label{fig:metr_circles}
\end{wrapfigure}

To assess the robustness of watermarking algorithms, it is common practice to not only evaluate detection accuracy metrics, but also to measure the resilience of watermark detection against attacks~\citep{zhu2018hidden, wen2023treering, fernandez2023stable}.
In the field of image watermarking, there are standard white-box attacks, as detailed in~\cite{peng2023intellectual, zhao2023invisible, wm_survey, zhao2023invisible}. These attacks involve transformations of the generated image to verify the robustness of the proposed method, and include operations such as rotation, JPEG compression, and cropping, followed by scaling, Gaussian blur, Gaussian noise, and color jitter. In our study, we also employ attacks utilizing generative models, such as the diffusion model~\cite{diffusion_ddpm, ddim} attack described in~\cite{zhao2023invisible}, as well as the VAE~\cite{vae, vae2018_attack} attack detailed in~\cite{zhao2023invisible}. The VAE attack involves embedding an image into the latent space of a Variational Autoencoder (VAE) and then reconstructing it. The diffusion model attack works by denoising injected Gaussian noise on the generated image with the aim of altering the image to erase the watermark.

\vspace{-0.4cm}
\section{Methods}
\label{sec:method}


In this section, we first provide a detailed description of \METR, the Message Enhanced \TR~\cite{wen2023treering} algorithm and its extension \METRpp. We then present an algorithm designed to select the best hyperparameters for optimal \METR performance in any use case.

\vspace{-0.3cm}
\subsection{METR}
\label{sec:metr}

Figure~\ref{fig:METR} presents the pipelines for watermark generation and detection using \METR. 
Similar to \TR~\cite{wen2023treering}, \METR operates with the noise or latent noise of an image.
The watermarking procedure modifies this noise through the corresponding Fourier~\cite{fft} space.
The resulting image can be generated using any diffusion model with the DDIM~\cite{ddim} sampling algorithm. 

In \METR, messages consist of binary sequences encoded using concentric circles with radii increasing from 1 to $R$, where $R$ is defined as the \textbf{watermark radius}.
This radius is a fixed hyperparameter representing the number of bits in the message. Therefore, it is possible to encode $2^R$ messages.
Each bit of the message is represented by a single circle, where the ones are assigned the value of $S$ and the zeros are given the value of $-S$.
$S$ is another crucial parameter that we call the \textbf{message scaler}.
Figure~\ref{fig:metr_circles} illustrates examples of concentric circles with both encrypted and decrypted messages.
Algorithm~\ref{alg:gen} details the pseudocode for generating an image with message encryption using the METR algorithm.




\begin{figure}[h]
    \centering
    \begin{minipage}[t]{0.47\textwidth}
        \small
        \begin{algorithm}[H]
        \caption{\METR image generation}
        \label{alg:gen}
        \KwIn{
            Scaler $S$,\
            radius $R$,\
            binary message $m$ (\eg user ID)
        }
        \KwOut{Generated image}
        $\bx_T\sim \mathcal{N}(0, I)$\;
        $\bx_T^{\mathcal{F}} \gets \text{Fourier}(\bx_T)$\;
        \For{$r=1$ to $R$}{
            $\text{mask}_r \gets $ circle of radius $r$\;
            \eIf{$m[r] = 1$}{
                $\bx_T^{\mathcal{F}}[\text{mask}_r] \gets S$\;
            }{
                $\bx_T^{\mathcal{F}}[\text{mask}_r] \gets -S$\;
            }
        }

        $\bx_T^\text{wm} \gets \text{Inverse Fourier}(\bx_T^{\mathcal{F}})$\;
        $\text{Image} \gets \text{DDIM Sampling}(\bx_T^\text{wm})$\;
        \Return Image
        \end{algorithm}
    \end{minipage}%
    \hfill
    \begin{minipage}[t]{0.47\textwidth}
        \small
        \begin{algorithm}[H]
        \caption{METR message detection}
        \label{alg:det}
        \KwIn{Image, radius $R$, $p_0$}
        \KwOut{Predicted watermark}

        $\bx'_T \gets \text{DDIM Inversion}(\text{Image})$\;
        $\bx_T'^{\mathcal{F}} \gets \text{Fourier}(\bx'_T)$\;

        \If{$F_{\chi^2}(z(\bx_T'^{\mathcal{F}}))<p_0$\tcp*[f]{see Equation~\ref{eq:p-value}}}{
            \Return \text{NO WM}
        }
        $m \gets []$\;
        \For{$r=1$ to $R$}{
            $\text{mask}_r \gets $ circle of radius $r$\;
            \eIf{$\bx_T'^{\mathcal{F}}[\text{mask}_r].mean() > 0$}{
                $m.append(1)$\;
            }{
                $m.append(0)$\;
            }
        }
        \Return $m$
        \end{algorithm}
    \end{minipage}
\end{figure}

The message can be decoded by reverting the image to its noise using DDIM Inversion, followed by a transformation of the inverted image into the Fourier space. Then, we determine whether the image was watermarked or not by evaluating the p-value using Equation~\ref{eq:p-value} and comparing it with a previously defined threshold $p_0$.


The decryption process for a binary message requires determining the sign of the value for each circle, obtained by averaging all values across the circle.


The corresponding pseudocode for message decryption is shown in Algorithm~\ref{alg:det}.

\begin{figure}[ht]
    \begin{subfigure}[htp]{0.22\textwidth}
        \centering
        \includegraphics[width=0.99\linewidth]{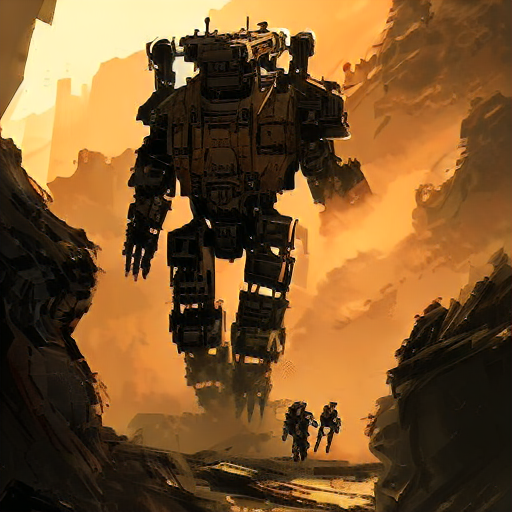}
        \subcaption{no WM}
    \end{subfigure}
    \hfill
    \begin{subfigure}[htp]{0.22\textwidth}
        \centering
        \includegraphics[width=0.99\linewidth]{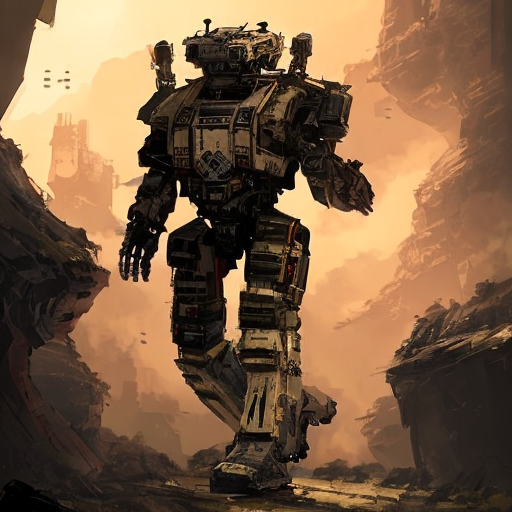}
        \subcaption{$r=4$}
    \end{subfigure}
    \hfill
    \begin{subfigure}[htp]{0.22\textwidth}
        \centering
        \includegraphics[width=0.99\linewidth]{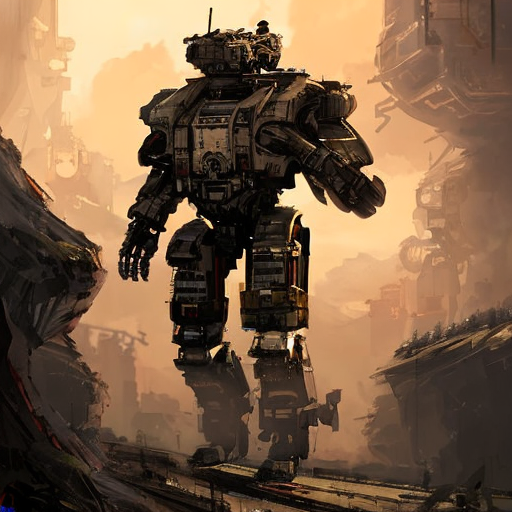}
        \subcaption{$r=16$}
    \end{subfigure}
    \hfill
    \begin{subfigure}[htp]{0.22\textwidth}
        \centering
        \includegraphics[width=0.99\linewidth]{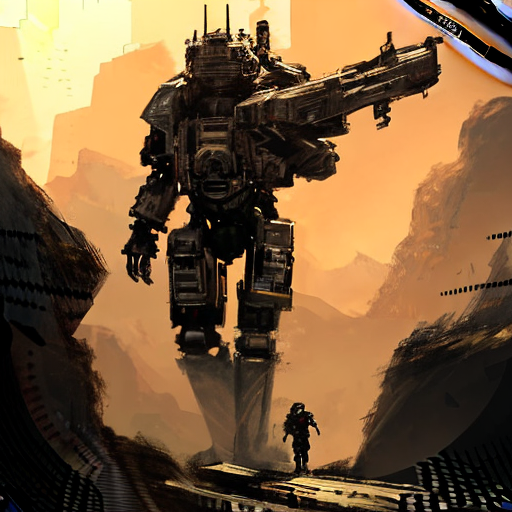}
        \subcaption{$r=34$}
    \end{subfigure}
\caption{Original image and the corresponding one with a \METR watermark, generated with a fixed scale $S$, but with different radii $r$.}
\label{fig:different_r}
\vspace*{-5mm}
\end{figure}
\begin{figure}[ht]
    \begin{subfigure}[htp]{0.22\textwidth}
        \centering
        \includegraphics[width=0.99\linewidth]{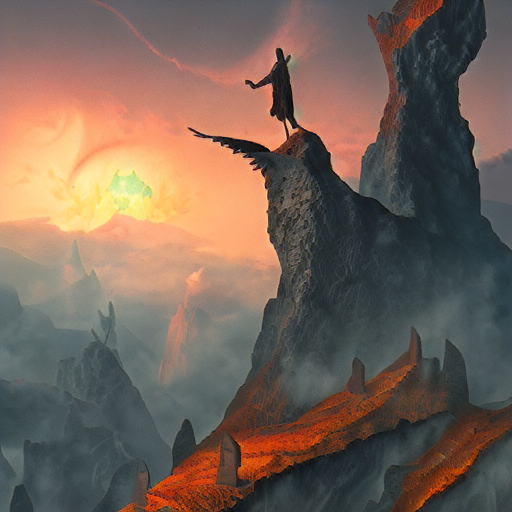}
        \subcaption{no WM}
    \end{subfigure}
    \hfill
    \begin{subfigure}[htp]{0.22\textwidth}
        \centering
        \includegraphics[width=0.99\linewidth]{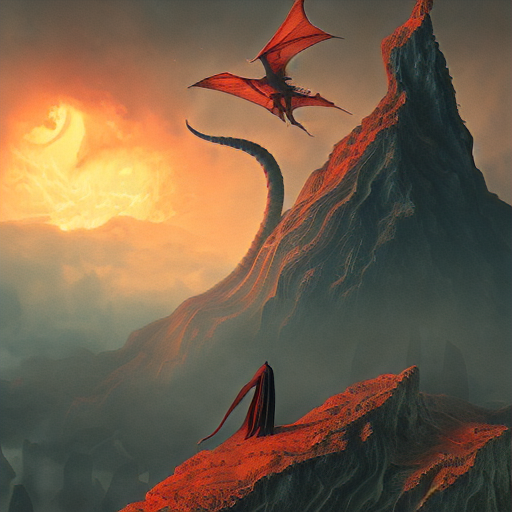}
        \subcaption{$S=60$}
    \end{subfigure}
    \hfill
    \begin{subfigure}[htp]{0.22\textwidth}
        \centering
        \includegraphics[width=0.99\linewidth]{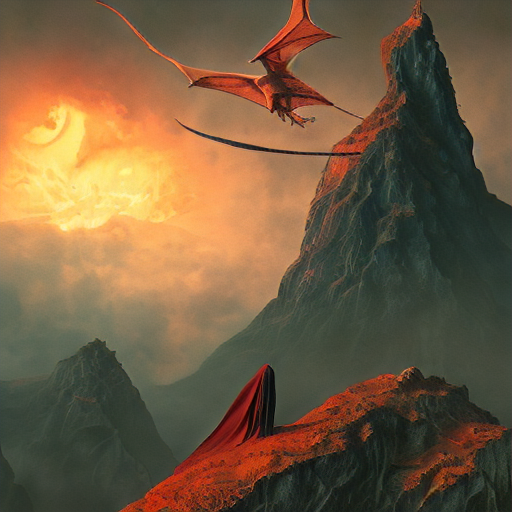}
        \subcaption{$S=100$}
    \end{subfigure}
    \hfill
    \begin{subfigure}[htp]{0.22\textwidth}
        \centering
        \includegraphics[width=0.99\linewidth]{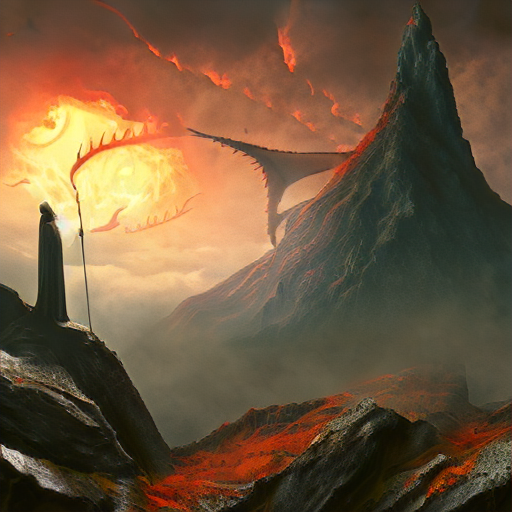}
        \subcaption{$S=140$}
    \end{subfigure}
\caption{Original image and the corresponding one with a \METR watermark, generated with fixed radius $r$, but with different scales $S$.}
\label{fig:different_s}
\vspace*{-5mm}
\end{figure}





\vspace{-0.3cm}
\subsection{Detection Resolution Metric}
\label{sec:det-resol}

In this section we describe how to select watermark radius $r$ and message scaler $S$, which are the parameters of \METR watermarking algorithm introduced in~Section~\ref{sec:metr}, and propose the metric Detection Resolution Metric for selecting the scale parameter.  

The increase of the radius parameter leads to the increase of the number of the potential messages, and the enlarging scale makes the message more detectable. However, increasing each of the parameter also leads to the production of corrupted images marked by various visible artifacts. See Figure~\ref{fig:different_r} and Figure~\ref{fig:different_s} for examples.


When aiming to encrypt the desired number of messages using the proposed algorithm, it is advisable to estimate the upper bound of possible messages for selecting the radius parameter $r$. Utilizing the smallest suitable radius is recommended in order to preserve the highest possible image quality.


When choosing the message scaler $S$, we recommend finding a balance to ensure precise watermark detection and high image quality within the proposed Detection Resolution metric. This metric is designed to capture the contrast between watermarked images and those without watermarks.
It is based on the detection distance~\cite{wen2023treering}, which is essentially an average error per pixel in the true watermark and restored watermark: $d_{\text {det}}(\bx_0) =
\frac{1}{|M|} \sum_{i = 1}^{|M|}
|\text{WM}_i-\text{Detect}(\bx_0)_i|$
here $M$ is the watermarked area, WM is a true watermark, $\bx_0$ is the original image with a possible watermark and \enquote{Detect} function is Fourier transform of DDIM Inversion of its argument.


\begin{figure}[!ht]
    \includegraphics[width=\linewidth]{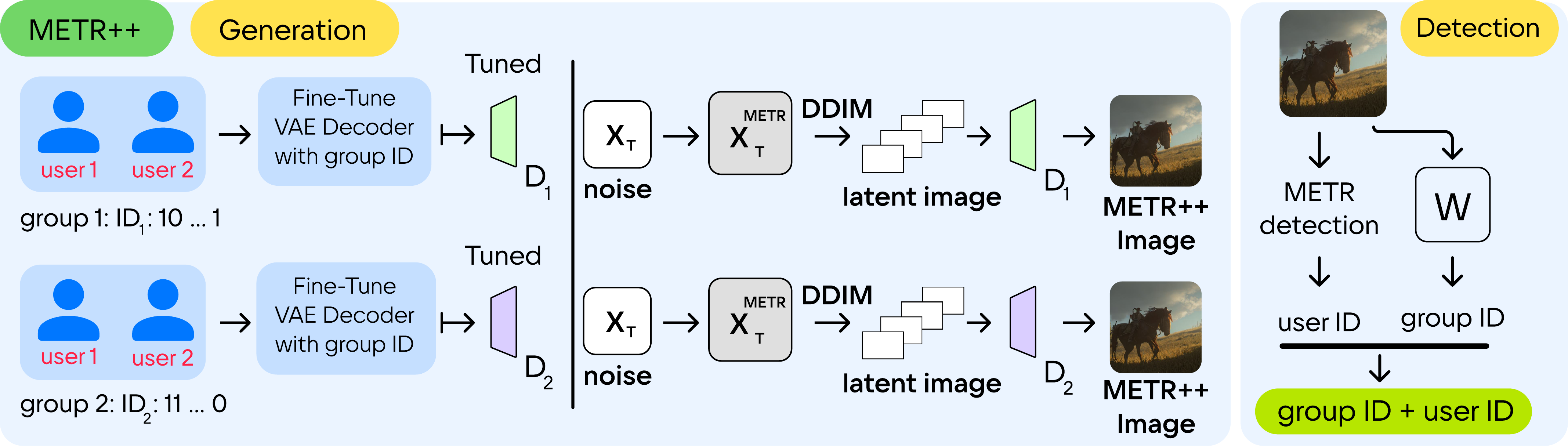}
    \caption{\METRpp pipeline where messages are divided into groups. \METR is used to encode messages inside a group, and \Stablesig~\cite{fernandez2023stable} is used to encode group itself.}
\label{fig:METR++}
\vspace*{-5mm}
\end{figure}


Detection Resolution metric computes the difference in detection distances between an image without a watermark $\bx_0$ and an image with one $\bx_0^*$: 
\begin{equation}\label{eq:det_res}
R_{\text {det}}(\bx_0, \bx_0^*)= d(\bx_0) - d(\bx_0^*)
\end{equation}

A binary message $m$ is considered accurately detected if the inverse process error does not negate the value of $S$. This means that $S - d(\bx_0^*) < 0$ should hold for watermarked images, and $S - d(\bx_0) > 0$ for non-watermarked ones. 
Furthermore, it is essential that the detection resolution metric, which is the error contrast between watermarked and non-watermarked images, is a relatively large value, compared to $S$. To specify this, we computed $g = R_\text{det} / S$ for situations with a perfect detection of message, and it occurred that $g=kS + b$, where $k$ and $b$ are linear fit parameters.
It means that to prevent detection collisions, we need $g\ge kS + b$ or $R_\text{det}/(kS^2 + bS) \ge 1$.
During early experiments, we found out that the most reliable $k$ and $b$ are $-2.23\cdot10^{-3}$ and $0.653$ respectively.




Since the error of the DDIM Inversion~\cite{ddim} process with a fixed model remains relatively constant with respect to the chosen image, we can assess the detection accuracy for a given value of $S$ using just a single pair of $(\bx_0, \bx_0^*)$. This way, we bring the selection of $S$ down to defining the possible range of its value and selecting the best one based on a subset of images. The algorithm consists of two steps. Initially, it measures the detection resolution using a generated image with and without watermark from the same noise. Then, if the criterion is satisfied, it assesses the image quality across the full test dataset. See Algorithm~\ref{alg:s-select} for details.

\begin{algorithm}[h]
\caption{Parameter $S$ search}
\label{alg:s-select}
\KwIn{
    latent noise $\bx_T\sim \mathcal{N}(0, I)$,
    test dataset $\mathbf{D} = \langle \text{prompt}, \bx_{\text{true}} \rangle$,
    image quality threshold $\Theta$,
    $k$, $b$, radius $r$
}
\KwOut{best message scaller $S$}


$\bx_T^{\mathcal{F}} \gets \text{Fourier}(\bx'_T)$\;
$p_\text{min},p_\text{max} \gets \min(\bx_T^{\mathcal{F}}), \max(\bx_T^{\mathcal{F}})$\;

\For{$S=p_\text{min}$ to $S=p_\text{max}$; step}{
    $\bx_0 \gets \text{Generate}(\bx_T)$\;
    $\bx_0^* \gets \text{Generate}(\text{METR}(\bx_T, S, r))$\;

    $d_0, d_0^* \gets d(x_0), d_(x_0^*)$\;
    Det\_Res $\gets \frac{R_\text{det}(d_0, d_0^*)}{kS^2 + bS}$\;

    \If{$d_0 > S$ and $d_0^* < S$ and Det\_Res $\geq 1$}{
        $D_{\text{test}} \gets$ Generate from prompts in $D$ with WM\;
        $L \gets \text{Eval}(D_{\text{test}}, D)$ \tcp*[f]{\ie FID}\;
        \If{$L < \Theta$}{\textbf{Return} S}
    }
}
\end{algorithm}

\vspace{-0.3cm}
\subsection{METR++}
\label{sec:metrpp}

Although \METR already provides support for encoding messages into images, the selected radius $r$ still restricts their number.
To overcome this, we developed \METRpp, an extension of the original \METR algorithm. \METRpp is designed to significantly increase the original algorithm's potential for encoding various unique messages, while being limited in terms of models that it can be applied to. The extended algorithm can only be used with Latent Diffusion~\cite{LDM}, which is currently the most commonly used image generation architecture~\cite{LDM, dalle, kandinsky}. \METRpp incorporates two watermarks into an image. As demonstrated in Figure~\ref{fig:METR++}, it adopts the \METR watermarking algorithm~\ref{sec:metr}, augmented by a VAE decoder derived from the latent diffusion model~\cite{LDM}, as proposed in \Stablesig~\cite{fernandez2023stable}.


To encode multiple messages, they are first categorized into groups with a capacity of $2^r$, a size that reflects the number of potential unique messages in the \METR watermarking algorithm. 
Then, each group is assigned a distinct ID, and uses a specifically fine-tuned VAE decoder designed to encode the group's ID as a \Stablesig watermark.
Consequently, \METRpp expands the capacity for encoding unique messages within \METR, multiplying it by the number of specially fine-tuned VAE decoders for each group.

The identification process consists of two parts, that can be done in parallel. First is the group's unique key decryption via the \Stablesig approach. Second is the \METR message decoding, which identifies a user within a group.
Decrypting the \METR message requires carrying out the inverse diffusion process on the image latent, that is obtained with a VAE encoder part of LDM, which weights stay intact.





To conclude, \METRpp can be adapted to any Latent Diffusion model and can encrypt approximately $2^r \cdot n$ unique messages, where $r$ is \METR's watermark radius, and $n$ is the number of fine-tuned VAE decoders. 




\vspace{-0.4cm}
\section{Experiments}
\label{sec:experiments}
In this section, we describe our experimental setup and present the results of experiments conducted using \METR and \METRpp.

\vspace{-0.3cm}
\subsection{Experimental Setup}
In all experiments, we utilize the base version of Stable Diffusion 2.1~\citep{LDM} and limit the generation process to 40 steps of DDIM~\cite{ddim} sampling.
The inverse diffusion process used for \METR message detection was run with no prompt for the same number of steps as the generation process.
The guidance scale was set to $7.5$.


For the baseline, we selected the Tree-Ring~\cite{wen2023treering} and Stable Signature~\cite{fernandez2023stable} watermarking algorithms for several reasons. In this paper, we aim to present a robust watermarking method capable of encrypting multiple messages. Tree-Ring~\cite{wen2023treering} is considered a state-of-the-art algorithm when it comes to robustness to various attacks~\cite{zhao2023invisible}. To our knowledge, Stable Signature is the only method capable of encrypting messages non-post-generation, where watermarking happens during the generation process~\cite{fernandez2023stable}. Finally, \METR and \METRpp build upon these existing watermarking algorithms.


\begin{wrapfigure}[10]{r}{0.45\linewidth}
    \centering
    \includegraphics[width=\linewidth]{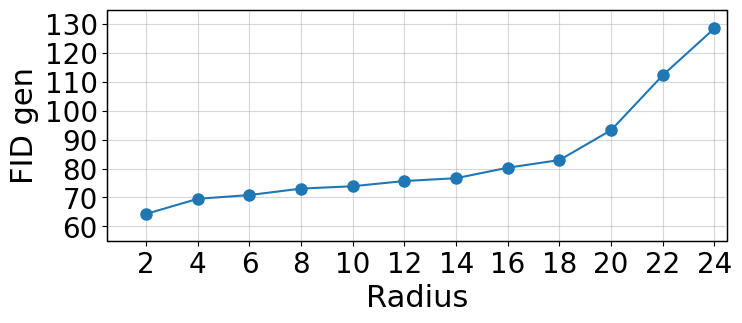}
    \vspace*{-8mm}
    \caption{\METR evaluation on image quality with different radii}
    \label{fig:fid_gen_r}
\end{wrapfigure}

We compare the models based on three main criteria: the accuracy of watermark detection, the accuracy of watermarked message decryption, and image quality, since we aim to ensure that the encrypted message does not degrade the generated image. 
The accuracy of watermark detection is measured by False Positive Rate (FPR), True Positive Rate (TPR), Area Under Curve (AUC) of Receiver Operator Characteristic (ROC), TPR value when FPR equals to 1\%, denoted as \enquote{TPR@1\%FPR}. The quality of message detection is assessed using Bit Accuracy, Word Accuracy (the accuracy of fully detected messages, as proposed in~\cite{zhao2023invisible}), and the detection resolution metric described in Section~\ref{sec:det-resol}. To assess image quality, we use \textbf{FID} (Fréchet Inception Distance)~\cite{fid} and \textbf{CLIP} score~\cite{clipscore}. FID is calculated by comparing the generated watermarked image to its corresponding ground-truth pair from the dataset, denoted as \textbf{FID gt}, or to a generated image created with the same prompt but without a watermark, denoted as \textbf{FID gen}. Comparison with the ground truth image indicates overall quality, while comparison with the generated image illustrates the impact of watermark encryption on the generation process. For the CLIP score, we measure the cosine similarity between the embedding of the watermarked image and the embedding of the reference prompt, following OpenCLIP-ViT/G~\cite{openclip}.

We utilized the MSCOCO-5000 dataset~\cite{mscoco} to evaluate \textbf{FID}. This dataset consists of 5000 paired images and prompts. To assess watermark detection accuracy, message detection quality, and image quality using \textbf{CLIP}, the images were generated using a subset of 1000 randomly selected prompts from the Stable Diffusion prompts~\cite{SD_prompts}.

\vspace{-0.3cm}
\subsection{METR Evaluation}
\label{sec: metr-rob}

In this subsection, we describe a set of experiments that compare \METR with \TR.
We focus on evaluating their robustness to white-box attacks, including ones that utilize generative models, as described in Section~\ref{sec:related-attacks}, watermark detection accuracy and the overall quality of the generated images.

To select proper \METR parameters, we first searched for the optimal message scale $S$ with the algorithm described in Section~\ref{sec:det-resol} in range $60 \leq S \leq 160$ with multiple different radii. The resulting optimal value was always between $80$ and $100$.
For the rest of our experiments, $S$ was set to $100$ unless specified otherwise. Regarding message radius $r$, we evaluated image quality for different radii on a subset of the MSCOCO-5000 dataset with 500 images. See Figure~\ref{fig:fid_gen_r} for results.

It is clear that keeping the radius as small as possible benefits the quality of the resulting image. However, we argue that increasing the radius to 16 can still maintain acceptable overall quality.
The benefit of doing so is that, this way, \METR can encode up to $2^{16}=65536$ messages.
In our experiments, we set $r=10$, which allowed us to encode $1024$ messages with almost no loss in quality.


\vspace{-0.3cm}
\begin{table*}[ht]
\centering
\caption{
    Detection metrics for \TR and \METR. $R_\text{det}$ denotes detection resolution metric. \enquote{B acc} and \enquote{W acc} represent Bit accuracy and Word accuracy, respectively. Note that for most of the white-box attacks \METR is as resilient as, or even more resilient than, \TR.
}
\renewcommand{\arraystretch}{1.5}
\begin{tabular}{c|cc||ccc}
\hline
\multirow{2}{*}{Attack} & \TR & \METR & \multicolumn{3}{c}\METR  \\
                        & \textbf{AUC} (\up)      & \textbf{AUC} (\up)  & \textbf{B acc} (\up) & \textbf{W acc} (\up) & $\mathbf{R_\text{det}}$ (\up)  \\ 
\hline
No attack               & 1.000 (+0.0\%)     & 1.000 (+0.0\%) & 0.991   & 0.910    & 42.7      \\
VAE~\cite{vae2018_attack}~$q=1$         & 0.999 (-0.1\%)     & 0.999 (-0.1\%) & 0.968   & 0.747    & 22.4      \\
diff,~150 steps                & 0.952  (-4.8\%)   & \bf{0.999}(-0.1\%)  & 0.967 (-3.3\%)    & 0.732    & 21.3      \\
rotate 75               & \bf{0.959} (-4.1\%)    & 0.907 (-9.3\%) & 0.694   & 0.033    & 5.1       \\
brightness 6.0          & 0.994 (-0.6\%)    & \bf{0.995} (-0.5\%) & 0.912   & 0.463    & 26.0      \\
noise $\sigma=0.1$           & \bf{0.941}  (-5.9\%)   & 0.833 (-16.7\%) & 0.695   & 0.134    & 7.9       \\
blur $r=4$                & 1.000  (-0.0\%)   & 1.000 (-0.0\%) & 0.979   & 0.805    & 34.1      \\
crop 0.75               & \bf{0.911}  (-8.9\%)   & 0.807 (-19.3\%) & 0.668   & 0.020    & 5.4       \\
JPEG 25                 & \bf{0.995} (-0.5\%)    & 0.992 (-0.8\%) & 0.953   & 0.719    & 27.0      \\
\hline
\end{tabular}
\label{fig:all-att-default}
\end{table*}

\textbf{Detection accuracy under image transformation attacks.} The detection metrics for cases with white-box attacks~\cite{zhao2023invisible} involving \METR and \TR are presented in Table~\ref{fig:all-att-default}. \enquote{B acc} and \enquote{W acc} represent Bit Accuracy and Word Accuracy, respectively. Since \TR cannot encode messages, these metrics are only calculated for \METR. As demonstrated by the results in Table~\ref{fig:all-att-default}, \METR is as resilient as, or even more resilient than, \TR against most image transformations. Both algorithms find rotation and cropping to be the most challenging types of attacks.

\begin{table}[!ht]
\centering
\caption{
   Message decryption accuracy for \METRpp and its components: \METR and \Stablesig watermarks. Note that the word accuracy of \METRpp is constrained by the lower word accuracy of either the \METR or \Stablesig watermarks.
}
\renewcommand{\arraystretch}{1.5}
\setlength{\tabcolsep}{2pt}
\begin{tabular}{c|cc|cc|cc} 
\hline
\multirow{2}{*}{Attack} & \multicolumn{2}{c|}{METR: ft VAE} & \multicolumn{2}{c|}{Stable-Signature} & \multicolumn{2}{c}{METR++}  \\
                        &  Bit acc & Word acc                         & Bit acc & Word acc                    & Bit acc & Word acc          \\ 
\hline
None                                                        & 0.991   & \underline{0.914}   & 0.995   & 0.843                    & 0.995   & 0.772            \\
VAE~~$q=1$                                                       & 0.970   & \underline{0.760}    & 0.490   & 0.000                      & 0.572   & 0.000         \\
diff,~150 steps                                             & 0.966   & \underline{0.719}    & 0.477   & 0.000                  & 0.561   & 0.000          \\
rot 75                                                        & 0.687   & \underline{0.029}    & 0.547   & 0.000                       & 0.572   & 0.000           \\
bright 6.0                                                   & 0.909   & \underline{0.460}    & 0.904   & 0.264                      & 0.905   & 0.209           \\
noise $\sigma=0.1$                                                & 0.694   & \underline{0.137}    & 0.539   & 0.000                       & 0.565   & 0.000          \\
blur $r=4$                                                       & 0.981   & \underline{0.816}    & 0.419   & 0.000                      & 0.516   & 0.000         \\
crop 0.75                                                        & 0.669   & 0.023    & 0.982   & \underline{0.569}                       & 0.928   & 0.013          \\
JPEG 25                                                          & 0.941   & \underline{0.672}    & 0.769   & 0.000                       & 0.799   & 0.000         \\
\hline
\end{tabular}
\label{fig:all-att-metrpp}
\end{table}

\begin{figure}[ht]
\centering
    \begin{subfigure}[htp]{0.44\linewidth}
        \centering
        \includegraphics[width=\linewidth]{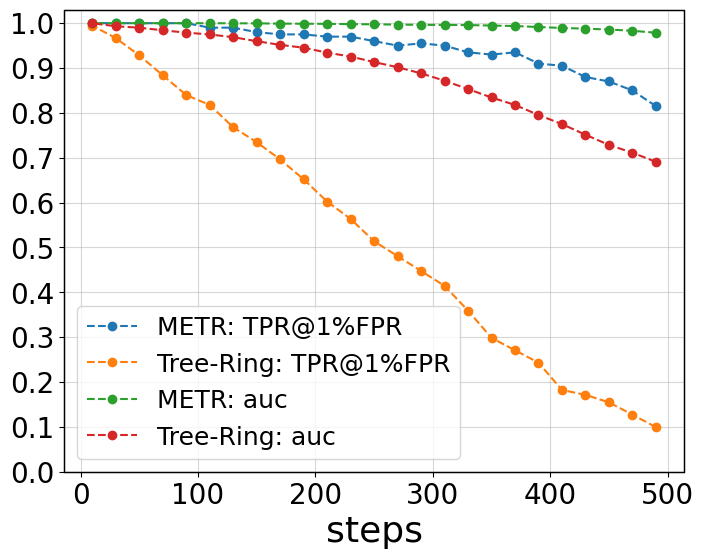}
        \subcaption{Watermark detection}
    \end{subfigure}%
    \hspace{0.5cm}
    \begin{subfigure}[htp]{0.44\linewidth}
        \centering
        \includegraphics[width=\linewidth]{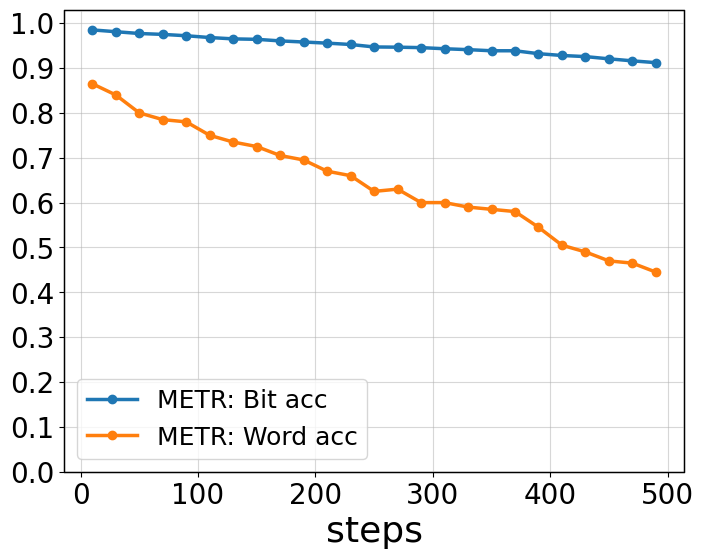}
        \subcaption{Watermark decryption}
    \end{subfigure}
    \caption{Watermark detection for \TR and \METR with the Diffusion attack~\cite{vae2018_attack}. Note that the watermark detection metrics for \METR are always higher than for \TR and decrease with the number of diffusion steps more slowly.}
    \label{fig:diffusion_attack}
    \vspace*{-5mm}
\end{figure}

\begin{figure}[ht]
\centering
    \begin{subfigure}[htp]{0.44\linewidth}
        \centering
        \includegraphics[width=\linewidth]{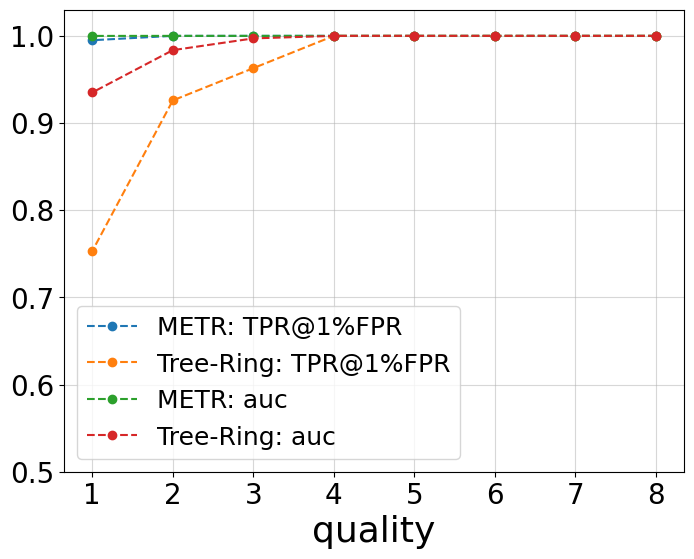}
        \subcaption{Watermark detection}
    \end{subfigure}%
    \hspace{0.5cm}
    \begin{subfigure}[htp]{0.44\linewidth}
        \centering
        \includegraphics[width=\linewidth]{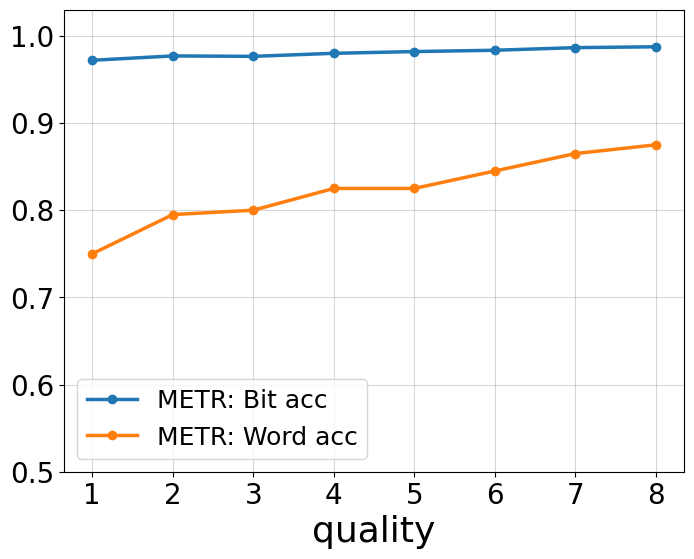}
        \subcaption{Watermark decryption}
    \end{subfigure}
    \caption{Robustness of \TR and \METR with a VAE attack~\cite{vae2018_attack}. Note that the detection metrics for \TR are below 1 until $q=4$, while for \METR they always equal 1.}
    \label{fig:vae_attack}
    \vspace*{-5mm}
\end{figure}

\textbf{Detection accuracy under generative models' attacks.} We performed a diffusion attack using the base version of Stable-Diffusion 2.1~\cite{LDM} and a VAE attack using the model \cite{vae2018_attack} with a quality hyperparameter $q=1$. The detection metrics for images generated with \TR and \METR watermark, both with the diffusion attack, can be seen in Figure~\ref{fig:diffusion_attack}. As one can see in the chart, the AUC and \text{TPR@1\%FPR} for METR are higher than those for \TR, and decrease with the number of diffusion steps more slowly. Similar results are shown for the VAE attack in Figure~\ref{fig:vae_attack}, parametrized by $q$. One can see that \TR detection metrics are below 1 until $q=4$, while \METR is robust to this attack and the watermark is almost always detected correctly. The \TR algorithm is not capable of message encryption, and thus word and bit accuracy are shown only for the \METR algorithm.


\textbf{Image quality.}  The results of comparing \METR with \TR are presented in Table~\ref{fig:quality-metr}. For the experiment, we sampled a random message for the \METR algorithm. It can be seen that the quality of images generated with the \METR watermark is close to those with the \TR watermark. In terms of the CLIP score, the performance of \TR is similar to that of an image without a watermark, while it drops by 2\% for \METR. When it comes to the FID score, the relative increase for images generated without a watermark and those with \TR and \METR watermarks are 1.7\% and 3.2\%, respectively.

\begin{table}[h]
\centering
\caption{
The image quality for images without watermarks is assessed using \TR and \METR. Note that for watermark-free images, those generated using \TR produce results that are nearly identical to those generated with \METR, although \TR scores are slightly higher.
}
\renewcommand{\arraystretch}{1.5}
\begin{tabular}{c|cccc} 
\hline
Method   & Message & \textbf{FID gen} (\down) &  \textbf{FID gt} (\down) & \textbf{CLIP} (\up)   \\ 
\hline
No WM     & - & -                & 25.570 (+0.0\%) & 0.364 (-0.0\%) \\ 
\hline
Tree-Ring & - & 10.283 (+0.0\%)  & 26.007 (+1.7\%) & 0.364 (-0.0\%) \\ 
\hline
METR      & + & 10.971 (+6.7\%)  & 26.398 (+3.2\%) & 0.357 (-1.9\%) \\
\hline
METR++      & + & 11.017 (+6.9\%)  & 25.206 (-1.0\%) & 0.365 (+0.3\%) \\
\hline
\end{tabular}
\label{fig:quality-metr}
\end{table}

In conclusion, the \METR algorithm demonstrates high resilience to all white-box attacks. The accuracy of watermark detection, even without any attacks or under white-box attack conditions, is very close to the detection accuracy achieved by \TR, and surpasses \TR for attacks with generative models~\cite{zhao2023invisible}. As for image quality, we noticed only a slight decrease with the \METR algorithm when compared to \TR, with less than a 2\% difference. However, \METR makes it possible to encode multiple unique messages in a watermark. The accuracy of the decrypted messages showed high resilience to the majority of white-box attacks, with exceptions being rotation and cropping. The word accuracy for messages without any attack reached 0.91.

\vspace{-0.3cm}
\subsection{METR++ Evaluation}

In this section, we perform a series of experiments to evaluate the robustness of \METRpp against white-box attacks. We also compare its detection metrics with those of \METR and the \Stablesig watermark~\cite{fernandez2023stable}.

As previously detailed in Section~\ref{sec:metrpp}, METR++ consists of the METR watermark and a fine-tuned VAE decoder designed to encrypt a 48-bit \Stablesig~\cite{fernandez2023stable} message. Therefore, our evaluation begins by investigating whether the inclusion of the METR watermark in the Fourier space of the latent noise decreases detection resilience.
To assess this, we fine-tune the VAE decoder using images drawn from standard latent noise and a distribution in which the METR watermark is embedded into the latent noise space.
We decode the messages encoded into the images using both the standard VAE decoder and the one that was fine-tuned on images with the METR watermark.
The results are presented in Table~\ref{fig:vae-tuning}.
The term \nrm-METR refers to the VAE decoder that was fine-tuned on images sampled from a normal distribution and subsequently used to work with images with the METR watermark.
It can be observed that the bit accuracy of the decrypted \Stablesig message in \METRpp is not affected by whether the VAE decoder has been fine-tuned on images sampled from latent noise with the embedded METR watermark. When it comes to the attacks, the bit accuracy of the decrypted \Stablesig message remains almost constant, showing a change of less than 1\% compared to basic \Stablesig.

\vspace{-0.3cm}
\begin{table}[h]
\centering
\caption{
    The bit accuracy of the Stable-Signature watermark. Note that the bit accuracy of the decrypted \Stablesig message from the images containing a \METR watermark remains nearly constant, with a variation of less than 1\% compared to the images without a \METR watermark.
}
\renewcommand{\arraystretch}{1.5}
\begin{tabular}{c|cccc}
\hline
VAE decoder      & no attack & crop 0.1 & bright 2 & jpeg 50 \\
\hline
\nrm-\nrm & 0.997      & 0.969  (+0.0\%)  & 0.990 (+0.0\%)   & 0.870 (+0.0\%)  \\
\hline
\nrm-METR                & 0.997      & 0.972 (+0.3\%)   & 0.991 (+0.1\%)   & 0.867 (-0.3\%)  \\
\hline
METR-\nrm                & 0.997      & 0.971  (+0.3\%)  & 0.989 (-0.1\%)    & 0.866 (-0.4\%)   \\
\hline
METR-METR                               & 0.997    & 0.974 (+0.5\%)    & 0.990 (-0.0\%)   & 0.863   (-0.8\%) \\
\hline
\end{tabular}
\label{fig:vae-tuning}
\end{table} 


The detection accuracy of \METRpp consists of the detection accuracies for both the \Stablesig and \METR watermarks. Table~\ref{fig:all-att-metrpp} presents the results of the detection accuracy evaluation.
The term \enquote{METR: ft VAE} refers to the accuracy of detecting the METR watermark when processed through the \METRpp pipeline.
\enquote{Stable-Signature} refers to the accuracy of decrypting the referenced watermark using the \METRpp pipeline.
Lastly, \enquote{\METRpp} denotes the overall detection accuracy of the complete \METRpp message, which includes both the \METR and \Stablesig messages.
The detection accuracy of the \METR watermark remains unaffected by the \METRpp pipeline, as the image decoded by the fine-tuned decoder closely resembles the one generated by the original VAE decoder. Consequently, both the Bit and Word accuracy for \METR watermark detection, as previously shown in Table~\ref{fig:all-att-default}, demonstrate robustness against most white-box attacks, aside from rotation and cropping. On the other hand, \Stablesig is vulnerable to white-box attacks, which in turn impacts the Word Accuracy of \METRpp. In Table~\ref{fig:all-att-default}, we highlighted a higher Word Accuracy between the \METR and \Stablesig watermarks. It is important to note that the robustness of \METRpp, in terms of word accuracy, is limited to the lesser robustness of the two watermarks since it requires the correct decryption of both and in terms of bit accuracy it is highly correlated with \Stablesig part, because it has more bits, than \METR part of \METRpp.

In terms of the potential number of messages, we use 58 bits for each message therefore, it is possible to encode  $2^{58}$ unique messages. When considering the practical application of either \METR or \METRpp, it is important to evaluate the trade-off between the capability to encode numerous unique messages and the resilience of the watermarking method against white-box attacks.


\vspace{-0.4cm}
\section{Future Work}
In this study, we assessed the detection accuracy of \METR and \METRpp against several white-box attacks. \METR demonstrated high resilience to most of the attacks, including state-of-the-art ones. We also propose researching methods to fine-tune the weights of the generative model to enhance the robustness of the \METR watermarking algorithm. When it comes to \METRpp, this method showed low robustness to most attacks due to its word accuracy robustness being constrained by the lowest detection accuracy among the two messages. The \Stablesig watermark message is generally less robust to attacks than the \METR watermark message. Therefore, to increase the encoding capacity of the \METR watermarking algorithm, we suggest exploring alternative watermarking methods that could replace \Stablesig, or investigating \Stablesig modifications within \METRpp that improve its detection accuracy under attacks.

Another area of research that needs further exploration is the impact of specific messages on the quality of the generated images. Our findings indicate that the quality of images generated using an average \METR message is comparable to those produced with the \TR watermark. However, in real-world applications, it is essential to examine various combinations of binary values to ascertain whether certain messages with \enquote{extreme values} might significantly affect image quality. This consideration could be crucial for commercial projects aiming to ensure consistent image generation quality for all their users.



\vspace{-0.4cm}
\section{Conclusion}
In this paper, we introduced the \METR watermarking algorithm, which can be applied to any diffusion model architecture with a non-probabilistic sampler (\eg DDIM~\cite{ddim}). Building upon the \TR watermarking algorithm, \METR retains its robustness in detection under white-box attacks and high image quality. The most significant advantage of \METR is its ability to encrypt multiple unique messages without the need for model's weights fine-tuning. This positions the proposed watermarking algorithm as one that can be considered the current leading watermarking algorithm in terms of robustness, image quality, and message encoding capacity. We also propose an extension of \METR, named \METRpp, which is specifically tailored to Latent Diffusion Models and requires additional fine-tuning of a VAE decoder for each new user group. \METRpp increases the potential number of encoded messages by the factor of fine-tuned VAE decoders. Comparing \METR with its extension \METRpp shows that the decision of which algorithm to use for practical applications should balance the total number of messages that can be encoded and the watermark's overall robustness to attacks. Implementation of our work can be found at \url{https://github.com/deepvk/metr}.

\bibliography{mybibfile}

\end{document}